\documentclass[letterpaper, 10 pt, journal, final]{IEEEtran} 

\IEEEoverridecommandlockouts                              

\usepackage{graphicx}
\usepackage{caption}
\usepackage{subcaption}
\usepackage{makecell}
\usepackage{tikz}
\usepackage{pgfplots}
\pgfplotsset{compat=1.17}

\usepackage{standalone}
\usepackage{multirow}
\usepackage{comment}
\usepackage{enumitem}
\usepackage{amsmath}
\usepackage{amssymb}
\usepackage{amsfonts}

\usepackage{multicol}

\usepackage{hyperref}
\hypersetup{
    colorlinks=true,
    linkcolor=blue,
    filecolor=magenta,      
    urlcolor=cyan,
    pdftitle={Overleaf Example},
    pdfpagemode=FullScreen,
    }

\newcommand\X[0]{Ev-Conv}

\newcommand\draftmod[1]{\noindent{\color{black}{#1}}}

\newcommand\draftmodadd[1]{\noindent{\color{black}{#1}}}

\newcommand\addition[1]{\noindent{\color{black}{#1}}}
\newcommand\remove[1]{\noindent{\color{red}{}}}

\begin{document}

\title{\LARGE \bf
\X{}: Fast CNN Inference on Event Camera Inputs\\For High-Speed Robot Perception}

\author{Sankeerth Durvasula$^{1}$, \IEEEmembership{Student Member IEEE,} Yushi Guan$^{1}$ and Nandita Vijaykumar,$^{1}$\IEEEmembership{Member IEEE}
\thanks{Manuscript received: September 22nd, 2022; Revised January 23rd 2023; Accepted February 21st 2023.}
\thanks{This paper was recommended for publication by Editor Markus Vincze upon evaluation of the Associate Editor and Reviewers' comments.} 
\thanks{$^{1}$Sankeerth Durvasula, Yushi Guan and Nandita Vijaykumar are with the University of Toronto, Toronto, ON, Canada 
        {\tt\footnotesize sankeerth.durvasula@mail.utoronto.ca; }}%
\thanks{Digital Object Identifier (DOI): see top of this page.}
}

\markboth{IEEE Robotics and Automation Letters. Preprint Version. Accepted February, 2023}
{Durvasula \MakeLowercase{\textit{et al.}}: \X{}: Fast CNN Inference on Event Camera Inputs\\For High-Speed Robot Perception} 

\maketitle
\thispagestyle{empty}
\pagestyle{empty}


\begin{abstract}
Event cameras capture visual information with a high temporal resolution and a wide dynamic range. This enables capturing visual information at fine time granularities (e.g., microseconds) in rapidly changing environments. This makes event cameras highly useful for high-speed robotics tasks involving rapid motion, such as high-speed perception, object tracking, and control. However, convolutional neural network inference on event camera streams cannot currently perform real-time inference at the high speeds at which event cameras operate{\textemdash}current CNN inference times are typically closer in order of magnitude to the frame rates of regular frame-based cameras. Real-time inference at event camera rates is necessary to fully leverage the high frequency and high temporal resolution that event cameras offer. This paper presents \X{}, a new approach to enable fast inference on CNNs for inputs from event cameras. We observe that consecutive inputs to the CNN from an event camera have only \emph{small differences} between them. Thus, we propose to perform inference on the difference between consecutive input tensors, or the \emph{increment}. This enables a significant reduction in the number of floating-point operations required (and thus the inference latency) because \emph{increments} are very sparse. We design \X{} to leverage the irregular sparsity in increments from event cameras and to retain the sparsity of these increments across all layers of the network. We demonstrate a reduction in the number of floating operations required in the forward pass by up to $98\%$. We also demonstrate a speedup of up to $1.6\times$ for inference using CNNs for tasks such as depth estimation, object recognition, and optical flow estimation, with almost no loss in accuracy. 
\end{abstract}

Code: \url{https://github.com/utcsz/evconv}

\IEEEpeerreviewmaketitle

\section{INTRODUCTION}
\draftmod{Event-based cameras have emerged as a promising method to generate visual information for robot vision as they capture high-speed changes, are resilient to motion blur, have high dynamic range, and consume low power. As a result of the high frequencies at which event cameras capture changes in the environment, they are useful for several robotics tasks which require fast perception of the environment. For example, event cameras are used to detect and dodge fast-moving objects on a quadrotor with fast independent object motion estimation~\cite{evimo, evdodgenet} and high-speed 3D reconstruction on scenes with fast-moving objects~\cite{mc3d}. Event cameras are composed of a set of pixels that register changes in the intensity of light falling on them. These changes are then streamed asynchronously as a stream of packets of the form~$(u_x, u_y, t, p)$, containing the pixel coordinate~$(u_x, u_y)$, the timestamp~$t$ at which the intensity change occured, and the polarity~$p=+1/-1$, corresponding to an increase or decrease in intensity. \draftmod{In contrast to frame-based cameras, which capture frames at a rate of 30 or 60 frames per second, event cameras stream packets at a rate of 1-10 MHz, which allows them to detect any environmental changes almost instantly.}

Convolution neural networks have shown state-of-art \draftmod{accuracy} in a number of vision tasks on inputs from event cameras, such as image classification~\cite{hats}, object detection~\cite{evobjmo, asynchronous_sparse_cnn}, human pose estimation~\cite{evdataset_pose, ev_neuropose}, depth estimation~\cite{rpg_e2depth, ramnet, aegnn}, optical flow estimation~\cite{evflownet, continuous_optical_flow}, independent object motion estimation~\cite{evimo,evdodgenet, evobjmo} and semantic segmentation~\cite{ev_semantic_segmentation, ev_ess}.} A fundamental challenge with using CNN inference on event camera inputs is that it requires high latencies. \draftmod{The processing time of CNN inference makes its use challenging with high-speed real-time applications.} For example, a forward pass for end-to-end inference on EVFlowNet takes $30ms$ on a GTX 1050 GPU~\cite{evflownet} \draftmodadd{and 9.5ms on a desktop RTX 3060}. High inference latencies negate the benefits of using event cameras. 

As a result of event cameras capturing changes in intensities at high frequencies, the changes between consecutive input features to the CNN are often small. However, processing each input still requires the full set of dense computations during CNN inference. Existing methods for CNN inference on event camera inputs treat each input as an independent image that requires end-to-end dense tensor computations, incurring large inference latency. 
We propose \X{}, a new approach to accelerate CNN inference for event camera inputs where we perform inference on the difference between consecutive input tensors (referred to as \emph{increment tensors}) instead of the input tensors themselves. We leverage the sparsity in increment tensors to accelerate CNN inference by skipping unnecessary computations on zero values.

We identify \draftmod{two} major challenges in effectively leveraging the sparsity of the increment tensors for event camera inputs. First, the sparsity in increment tensors tends to be highly irregular, as the pixels that register changes tend to be more scattered. Existing methods to enable speedup on sparse increment tensors exploit the property that large portions of the scene remain the same across frames when the camera isn't moving, leading to the sparsity of increment tensors being regular~\cite{cbinfer, deltacnn, skipconv, dynconv}. \draftmodadd{We introduce a sparse convolution operation to leverage this sparsity for speedup. As GPUs are optimized for dense computations, we introduce an approach to skip computations of convolution filters on a block of the tensor when all elements of the block are $0$.} 
\draftmodadd{Second, the sparsity level significantly reduces at the deeper layers of the network for the following reasons: (i) Convolution and matrix multiply operations spread the distribution of 0s in input tensors. We however observe that a large percentage of the tensor elements tends to be numerically small. We can significantly increase the sparsity of increment tensors by introducing a rounding-off mechanism that sets small elements to $0$. (ii) CNN architectures have downsampling operations to extract high-level semantics at low spatial resolution levels. Through downsampling, the tensors become denser as the spatial resolution decreases.} To address this problem, we propose a technique called \emph{delayed integration} to preserve sparsity of increment tensors. 
We test the idea with a proposed Delayed UNet architecture and demonstrate significant floating point operation reductions but with similar accuracy as the original CNN. 

The \draftmod{faster inference we obtain when using \X{} makes it more suitable for robotics vision applications that use CNNs to process inputs from event cameras.} We evaluate \X{} for a range of tasks that use event camera inputs on different types of CNN architectures, including depth estimation, optical flow estimation, and object recognition. We demonstrate that \X{} is able to reduce the number of floating operations by up to $98\%$ and provide a speedup of up to $1.6\times$, with similar accuracy as the original network.

\section{RELATED WORK}
\textbf{DNN Architectures for Event Cameras.} Recently proposed  DNN architectures~\cite{eventnet, fast_ir_ev,ev_1megapixel} enable faster inference on event camera streams by using smaller networks with fewer weights. Graph neural networks (GNNs)~\cite{aegnn}, for example, represent events as nodes in a graph and this representation requires less computation as fewer events are processed at each step.
AEGNN~\cite{aegnn} enables a 200-fold reduction in the number of floating point operations required in this manner. However, despite this, inference using GNNs is orders of magnitude slower than using CNNs with similar numbers of parameters~\cite{aegnn}, making them impractical for real-time inference in real-world tasks. EventNet~\cite{eventnet} uses a PointNet~\cite{pointnet}-like architecture to use a more efficient encoding of the event stream input. This encoding enables real-time inference, however, it discards the time stamps of the events. This results in lower accuracy and effectiveness of the network itself for various tasks~\cite{evflownet}. 

\textbf{Sparse DNN architectures.} Given the high frequency at which event camera streams capture changes in the environment, events are sparse in time and also in their locations in the scene being captured. Some encodings for event camera streams, such as event histograms~\cite{evhistogram} and event queues~\cite{eventqueue}, enable the input tensors to a CNN to be expressed as \emph{sparse tensors}. Some approaches~\cite{submspconv,torchsparse,spconv} exploit the sparsity in the input tensors to implement sparse convolutions with fewer floating-point computations compared to their dense counterparts. For example, for sparse event stream encodings, submanifold sparse convolutions~\cite{asynchronous_sparse_cnn} can reduce floating point operations by $10\times$. However, these sparse operations require irregular accesses to memory and despite significantly reducing the number of floating point operations, there is no reduction in overall inference latency~\cite{spconv,minkowski}. Other networks such as SBNet can leverage sparsity effectively to generate lower inference latencies. However, SBNet was designed for LiDAR point cloud inputs and event camera inputs do not have regularity in sparsity. 

\textbf{Accelerating inference on video streams.} Several recent works~\cite{cbinfer, deltacnn, skipconv, dynconv, cnnacceleratemlsys} leverage the similarity between consecutive frames in a video to accelerate inference. These methods leverage the typical case where large parts of the scene are static and unchanged across consecutive frames in the video. Thus convolution operations need to be performed for the non-static portions of the scene only. However, these approaches are not effective with event camera streams which are typically used in more dynamic environments where there is no guarantee of large static sections. The high temporal resolution detection of changes in intensity leads to more unstructured and irregular sparsity in the CNN inputs. DeltaCNN~\cite{deltacnn} also leverages the sparsity in \emph{deltas} between consecutive video frames. However, DeltaCNN is not effective with event camera inputs as the sparsity in the input increments is far more irregular. \draftmod{In addition, these works do not address the decrease in sparsity due to the upsampling in commonly used CNN architectures such as UNets.} We propose \X{}, an approach designed to leverage the irregular sparsity seen in event camera streams to generate low latency inference.

\vspace{-.5cm}

\addition{\section{METHOD}}
\subsection{Difference Between Consecutive Convolution Inputs}
An event camera measures the change in intensity of light falling on each pixel asynchronously. Every individual event contains the following information: $(u_x, u_y, t, p)$: pixel coordinate, timestamp, and polarity. CNN inference on event inputs at time $\tau$ encodes all the events occurring within a time window of $\Delta$ (typically $50$ milliseconds) before $\tau$. The events in this window are grouped into a tensor representation to be used as input to the neural network. Commonly used types of event encodings for CNNs include: \textit{event-voxels}~\cite{rpg_e2depth, asynchronous_sparse_cnn, ramnet}, \textit{event-count}~\cite{asynchronous_sparse_cnn} and \textit{most recent timestamp}~\cite{evflownet}. 
\remove{During inference, the time difference between the windows on consecutive event camera inputs is very small, unlike a convolution operation across images. Online inference on event camera inputs is therefore bottlenecked by the CNN forward pass processing time. The difference between event encodings between two consecutive inputs consists of values that have small magnitudes.}

\subsubsection{Increment tensors and increment layers}
\label{sec:incrtensors}
\draftmodadd{We define the \emph{increment} tensor as the difference between the tensor at the current inference step and the previous inference step}. For an input tensor $\mathbf{x}(n)$ to an operator of the CNN at inference step $n$, we define the increment \draftmodadd{tensor} of $\mathbf{x}(n)$ as $\mathbf{x}_{\uparrow}(n)$ as:
\begin{equation}
    \label{eq:0}
    \mathbf{x}_{\uparrow}(n) = \mathbf{x}(n) - \mathbf{x}(n-1)
\end{equation}

Alongside each layer in the original CNN, we implement an \emph{increment layer} that receives an input increment. A corresponding \emph{output increment tensor} $\mathbf{y}_{\uparrow}(n)$ is generated, which should be the difference between two consecutive outputs $\mathbf{y}(n)$ and $\mathbf{y}(n-1)$ of the layer. We aim to evaluate the increment in the CNN output by replacing each forward pass layer with its increment layer. The output of an operator in \X{} with increment layers is an increment tensor over the previous output. To compute the final output $\mathbf{y}$ after $m$ inference steps, we sum the previous $m$ increment outputs and the output tensor at time $n-m$:
\begin{equation}
    \label{eq:incr_sum}
    \mathbf{y}(n) = \mathbf{y}(n-m) + \sum^n_{i=n-m+1} \mathbf{y}_{\uparrow}(i)
\end{equation}

\subsubsection{Forward pass with increment tensors}
\label{sec:challenges}
Treating encoded event tensors as individual inputs (similarly to images) leads to many expensive dense tensor computations in the forward pass. However, at high inference frequencies (1kHz), there is a large degree of overlap between consecutive inputs. Thus, two consecutive time windows have similar input tensors $\mathbf{x}(n)$ and $\mathbf{x}(n-1)$. In other words, $\mathbf{x}(n)-\mathbf{x}(n-1)$ is sparse, and many elements have small absolute values. 
The input increment tensors must be sparse to reduce the number of floating-point operations performed by each increment layer, enabling us to perform sparse operations to obtain the output increment. We identify two challenges in performing inference with increment tensors: 
\begin{enumerate}[leftmargin=0pt,labelwidth=-10pt]
\item \textbf{Insufficient sparsity of increment tensors in deeper layers.}
We find that the sparsity of the increment tensors is irregular and decreases significantly towards deeper layers in the network. This is because convolution and matrix multiplication operations on sparse tensors do not necessarily produce sparse outputs and increase the spread of nonzero values, making it infeasible to leverage sparsity for faster execution.

\item \textbf{Drop in sparsity in encoder-decoder architectures on upsampling.} A number of CNNs used in computer vision have an encoder-decoder structure. The encoders encode the input feature tensor into a dense intermediate tensor. The subsequent upsampling operations in the decoder produce more dense tensors, making it infeasible to accelerate increment layers.
\end{enumerate}

\subsection{Sparsification layer}
\label{sec:sparsification_layer}
To address the insufficient sparsity problem mentioned in Section~\ref{sec:challenges}, we devise a mechanism to retain the sparsity in the increment tensors using a rounding-off mechanism. Typically, due to the similarity between the two consecutive inputs, the elements of the input increments to each operator are not only sparse but also have very small values. To further improve sparsification, we round off to zero all elements whose absolute values are smaller than a threshold. To perform this sparsification, we add a new layer, the \emph{sparsification layer}, at various points in the CNN.

At $n=0$, we round off the elements to the nearest multiple of a parameter $k$. Thus all elements whose absolute values are above the parameter $k$ are rounded off to zero. This can be expressed mathematically as:
\begin{equation}
 \label{eq:out0}
    \mathbf{y}_{\uparrow}(0) = k\left\lfloor 0.5+ \frac{\mathbf{x}_{\uparrow}(0)}{k} \right\rfloor
\end{equation}

The difference between $\mathbf{y}_{\uparrow}(0)$ and the $\mathbf{x}_{\uparrow}(0)$, or the residual resulting from the operation is stored in $\mathbf{\delta}(0)$ (Eqn~\ref{eq:delta0}).
\begin{equation}
 \label{eq:delta0}
    \mathbf{\delta}(0) = \mathbf{y}_{\uparrow}(0) - \mathbf{x}_{\uparrow}(0)
\end{equation}

This residual $\mathbf{\delta}(0)$ is the error between the elements in $\mathbf{y}_{\uparrow}(0)$ and the true increment $\mathbf{x}_{\uparrow}(0)$. On subsequent increments, this residual should be added to future inputs to correct the error in output tensor increments. Therefore, at $t=1$, we have: 
\begin{equation}
 \label{eq:x1}
    \mathbf{x}_{corrected}(1) = \mathbf{\delta}(0) + \mathbf{x}_{\uparrow}(1)
\end{equation}

This corrected output $\mathbf{y}_{corrected}$ tensor is now to be sparsified using the same mechanism as in Eqn~\ref{eq:out0}.  Therefore, at time $t$, we have the following update equations to update a sparse tensor $\mathbf{y}_{\uparrow}$ from an input increment $\mathbf{x}_{\uparrow}$:
\begin{equation}
\label{eq:out_corrected_t}
    \mathbf{x}_{corrected}(n) = \mathbf{\delta}(n-1) + \mathbf{x}_{\uparrow}(n)
\end{equation}
\begin{equation}
 \label{eq:out_incr_t}
    \mathbf{y}_{\uparrow}(n) = k\left\lfloor 0.5+ \frac{\mathbf{x}_{corrected}(n)}{k} \right\rfloor
\end{equation}
\begin{equation}
 \label{eq:delta_t}
    \mathbf{\delta}(n) = \mathbf{\delta}(n-1) + \mathbf{x}_{corrected}(n) - \mathbf{y}_{\uparrow}(n)
\end{equation}
\draftmodadd{Eqns~\ref{eq:out_corrected_t} and \ref{eq:out_incr_t} show the operations performed on increment tensor $x\uparrow$ to produce the increment tensor y↑ by rounding-off small values of the input to $0$. $\delta(n)$ stores the round-off error.}


This sequence of steps \remove{depicted in Figure~\ref{fig:sparse_layer},}ensures that the elements of $\delta$ remain small. The error due to rounding off in the output of our network varies proportionally with respect to the values in $\delta$. We define a hyperparameter, the thresholding parameter $t_p$, which we use to calculate $k$. The value of $k$ is computed as $k = t_p\|\mathbf{x}\|$,\draftmodadd{ where $x$ is the rolling average of the input to the sparsification layer and $\|\|$ is the $L_2$ norm.}

\begin{figure}[htb!]
    \centering
        \begin{subfigure}{0.5\textwidth}
            \centering
            \includegraphics[width=0.75\linewidth]{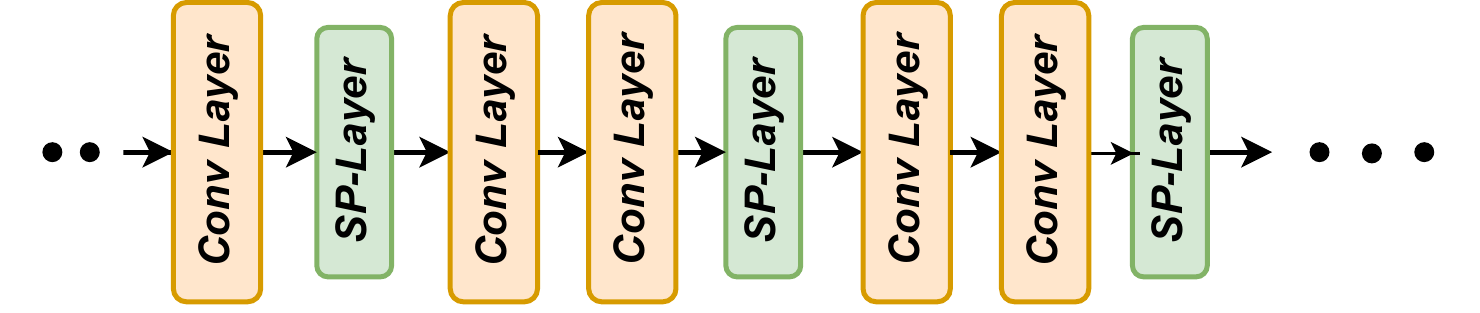}
            \caption{Sparsification layers (SP-Layers) at various points in YOLE}
            \label{fig:splayer_positioning}
        \end{subfigure}
        \hfill
        \begin{subfigure}{0.5\textwidth}
            \centering
            \includegraphics[width=0.75\linewidth]{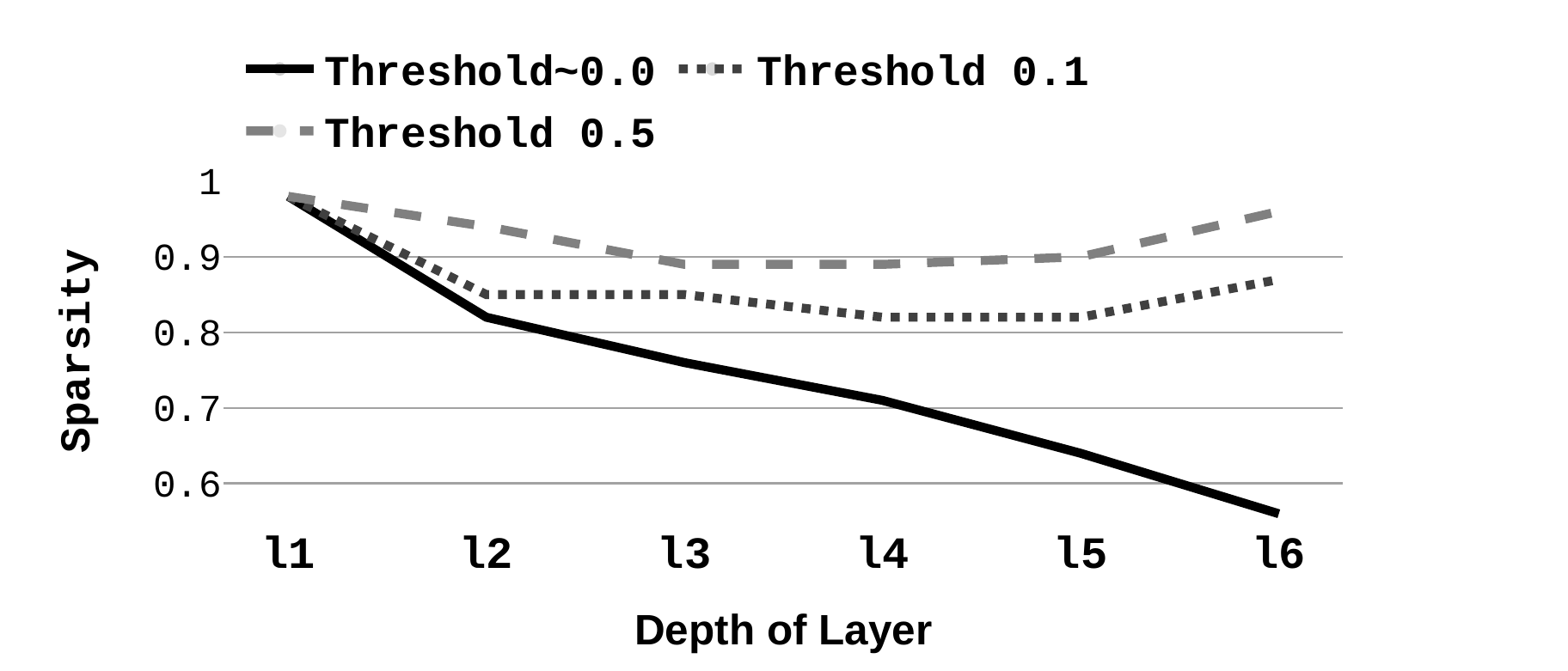}
            \caption{Sparsity in increments on YOLE on object detection}
            \label{fig:with_sparsification_layer}
        \end{subfigure}
        \caption{Sparsity of intermediate tensors with sparsification layers inserted at difference points of the network.}
        \vspace{-0.4cm}
        
\end{figure}

By adding sparsification layers before each pair of convolutional layers of our YOLE~\cite{yole} DNN for object detection (Figure~\ref{fig:splayer_positioning}), we can retain sparsity of the increment tensors at the deeper layers. Our mechanism ensures that each operator in the forward pass receives a sparse increment as an input, which results in a dramatic reduction in the number of floating-point operations required. We perform an experiment to measure the sparsity in the increment tensors in a YOLE~\cite{yole} network trained to detect objects on the Caltech101 dataset using increment layers. Figure~\ref{fig:with_sparsification_layer} depicts the sparsity in increment tensors for layers at various depths in the network. As depicted in the figure, before correction, there is a significant drop in sparsity towards the deeper layers of the network. 

\subsection{Mitigating Reduced Sparsity on Upsampling}
\draftmod{Many CNNs in computer vision, such as feature pyramid networks and UNets, have an encoder-decoder structure~\cite{Ronneberger2015UNetCN, Lin2017FeaturePN}. The encoder consists of a sequence of downsampling operations to create a high-level abstract feature that has lower spatial resolution than the input. This downsampling process leads to \emph{dense} intermediate tensors. In the decoder, the upsampling operations take these dense intermediate tensors as inputs and output more dense tensors. As a result, the sparse convolution operators do not improve inference speed on processing these dense tensors in the decoder.}

\begin{figure}[!htb]
    \vspace{-0.3cm}
    \centering
    \begin{subfigure}{0.45\textwidth}
    \centering
    \includegraphics[width=.7\linewidth]{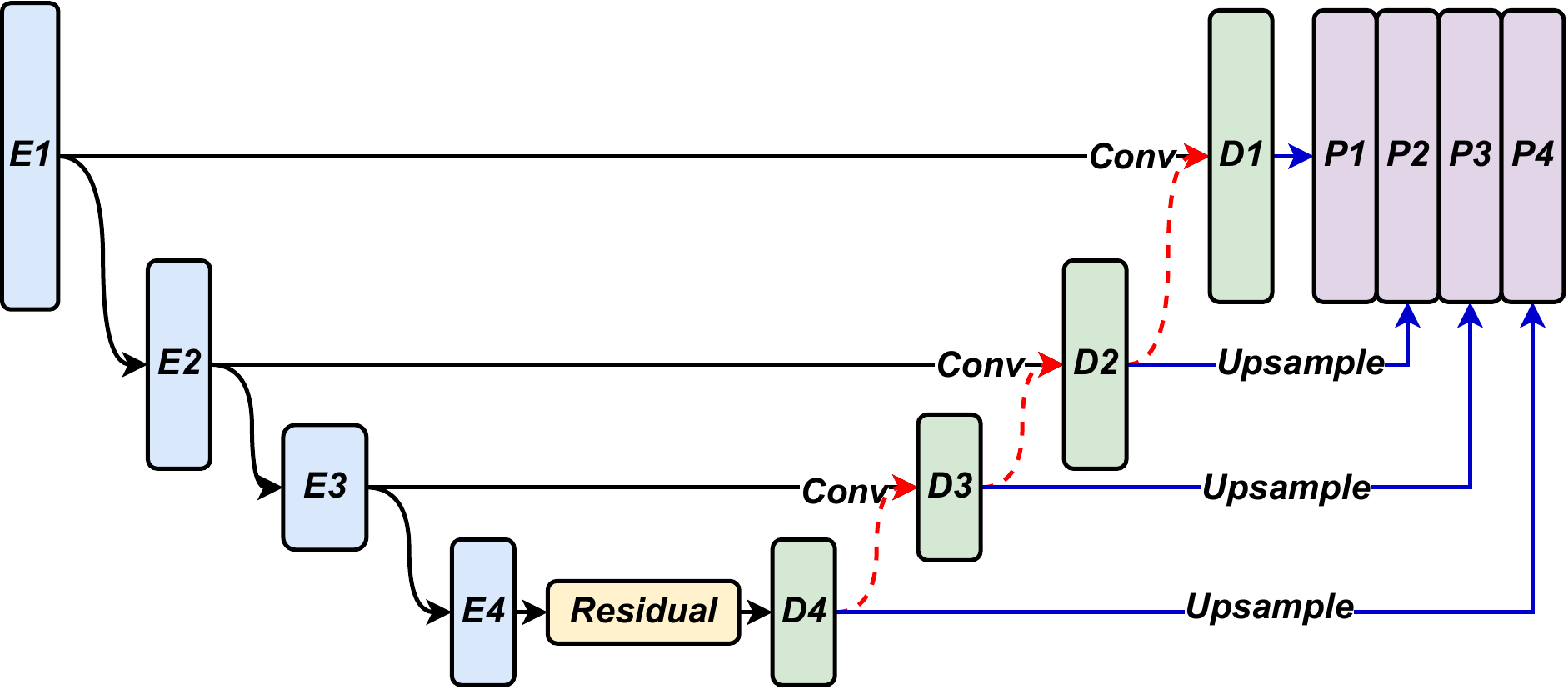}
    \caption{The dotted red line connections are removed from the original UNet architecture. $E$: encoder, $D$: decoder, $P$: prediction.}
    \label{fig:delayed_unet_architecture1}
    \end{subfigure}\\
    \begin{subfigure}{0.45\textwidth}
    \centering
    \includegraphics[width=0.7\linewidth]{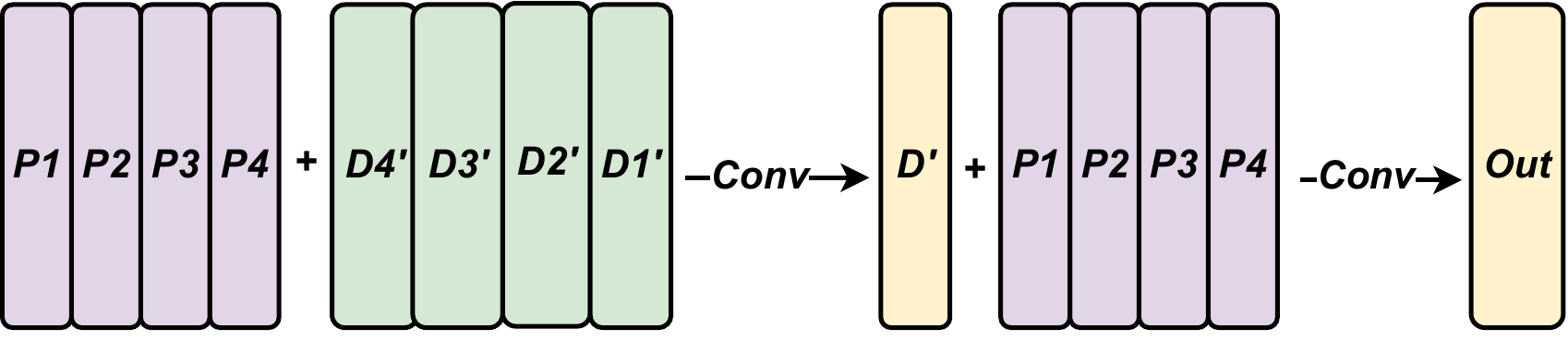}
    \caption{The upsampled decoder outputs $D'$ are concatenated with $P$ to generate the final prediction.}
    \label{fig:delayed_unet_architecture2}
    \end{subfigure}
    \caption{Delayed UNet architecture}
    \label{fig:delayed_unet_architecture}
    \vspace{-.9cm}

\end{figure}

In order to benefit from sparse convolution operation, we propose a \emph{delayed integration} technique and implement it in a delayed UNet architecture shown in Figures~\ref{fig:delayed_unet_architecture1} and \ref{fig:delayed_unet_architecture2}. In the original UNet, the output of an upsampling layer in the decoder is fed into the next decoder (labelled with dashed red arrows in Figure~\ref{fig:delayed_unet_architecture1}). In our proposed delayed UNet, the outputs of upsampling layers in the decoder are not fed into the subsequent levels of decoders. Instead, the upsampled outputs are concatenated with the upsampled predictions towards the end (thus the name \emph{delayed}). These concatenated tensors are then fed through two convolutions layers (Figure~\ref{fig:delayed_unet_architecture2}). Thus, the decoder layers do not receive the dense tensors from the lowest spatial resolution level and can benefit from sparse convolutions. We include the prediction explicitly as inputs to the final two convolutions since the predictions have been trained with auxiliary loss and reflect the network's prediction of optical flow at different spatial resolutions.

\subsection{Implementation of Increment Layers}
\label{sec:incrops}
\begin{figure}[htb!]
    \centering
    \includegraphics[width=0.60\linewidth, trim={0cm, 0cm, 0cm, 0cm}, clip]{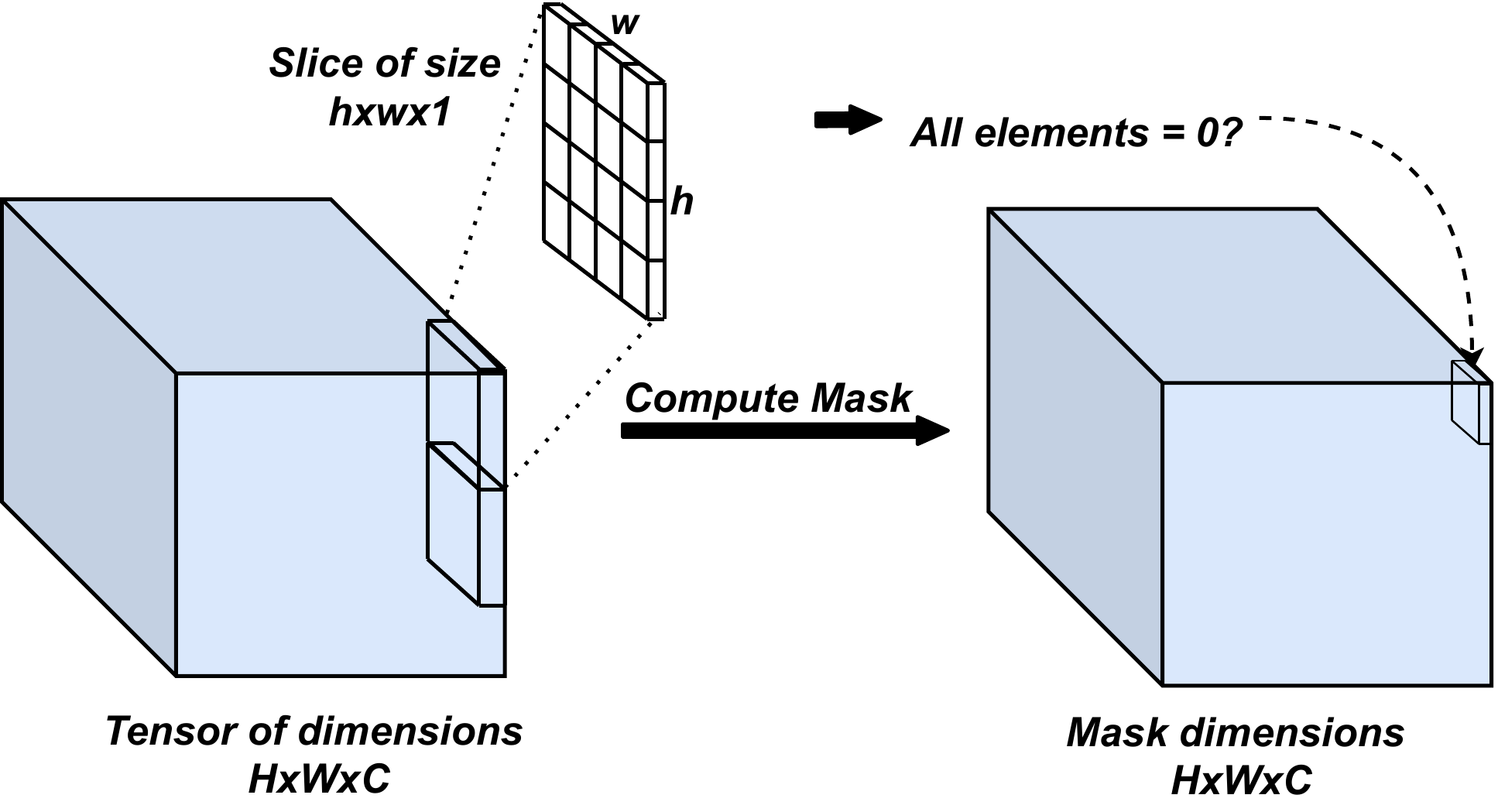}
    \caption{Compute mask on each 3D layout tensor}
    \label{fig:tensor_fscp}
    \vspace{-0.3cm}

\end{figure}

In this section, we describe the implementation of each of the increment layers corresponding to each operation for the forward pass. At inference step $n$, each operator takes as input an increment tensor~$\mathbf{x}_{\uparrow}(n)$ (as defined in Section~\ref{sec:incrtensors}) and a mask tensor~$\mathbf{x}_{m}(n)$. Each element of the mask indicates whether a region of the input tensor can be skipped for computation by the operator as it is entirely comprised of zeros. In CNNs where the input increment tensor and the mask tensor have dimensions $\mathbb{R}^{H\times W\times C}$ (i.e., height, weight, and channel), each element of the mask summarizes the sparsity of a smaller section of $\mathbb{R}^{h\times w\times1}$ elements in the input tensor, as depicted in Figure~\ref{fig:tensor_fscp}. For each operator, we estimate the following parameters: the output increment tensor $\mathbf{y}_{\uparrow}$ and the output mask $\mathbf{y}_{m}$. We drop the reference to a specific inference $n$ when there is no ambiguity.

We denote linear operations, including \textbf{convolution} and \textbf{matrix multiplication} operations, with $\theta_{L}$. For the increment tensor, we directly compute the increment in the output using increment in the input, as seen in Eqn~\ref{eq:linop1} and \ref{eq:linop2}.
\begin{gather}
\label{eq:linop1}
\resizebox{.89\hsize}{!}{$\theta_{L}(\mathbf{x}(n)) = \theta_{L}(\mathbf{x}(n-1) + \mathbf{x}_{\uparrow}(n))
= \theta_{L}(\mathbf{x}(n-1)) + \theta_{L}(\mathbf{x}_{\uparrow}(n))$} \\
\label{eq:linop2}
\mathbf{y}_{\uparrow}(n) = \theta_{L}(\mathbf{x}_{\uparrow}(n))
\end{gather}

We give an overview of our implementation of sparse convolution in Section~\ref{sec:conv_impl}.

\textbf{Addition.} We denote two inputs to an addition operation with subscripts 1 and 2 respectively. The output increment is implemented simply as a sum of two input increment tensors using a dense to sparse tensor addition (Eqn~\ref{eq:addincr}). This value is initialized at the beginning of the inference. The mask of the result of the addition of two sparse tensors can be computed with a bitwise OR operation ($\mid$) as shown in Eqn~\ref{eq:addmask}. 
\vspace{-1cm}
\begin{multicols}{2}
\begin{equation}
\label{eq:addincr}
\mathbf{y}_{\uparrow} = \mathbf{x}_{1,\uparrow} + \mathbf{x}_{2,\uparrow} 
\end{equation}\break
\begin{equation}
\label{eq:addmask}
\mathbf{y}_{m} =\mathbf{x}_{1,m}\mid \mathbf{x}_{2,m}
\end{equation}
\end{multicols}

\textbf{Activation.} \draftmodadd{Increment operators for non-linear functions like activations ($\theta_{a}$) is computed as the difference between the values of activation at step $n$ and activation computed at step $n-1$. We introduce an additional parameter $x_{acc}$ which is an estimate of the input in the previous run, calculated as the sum of the corresponding increment tensors until the current inference step. Using this, the difference between outputs of $\theta_a$ can be computed as shown in Eqn~\ref{eq:activation_incr}.  The input mask is equal to the output mask (Eqn~\ref{eq:activation_mask}). The estimate of the input to the operator is updated as shown in Eqn~\ref{eq:activation_accum}.}
\vspace{-.1cm}
\begin{gather}
\label{eq:activation_incr}
\mathbf{y}_{\uparrow} = \theta_{a}(\mathbf{x}_{\uparrow}+\mathbf{x}_{acc}) - \theta_{a}(\mathbf{x}_{acc})
\end{gather}
\vspace{-1.5cm}
\begin{multicols}{2}
  \begin{equation}
\label{eq:activation_mask}
\mathbf{y}_{m} = \mathbf{x}_{m}
  \end{equation}\break
  \begin{equation}
\label{eq:activation_accum}
\mathbf{x}_{acc} \gets \mathbf{x}_{acc} + \mathbf{x}_{\uparrow}
\end{equation}
\end{multicols}
\vspace{-.1cm}

\textbf{Elementwise Multiply.} Elementwise multiplication maintains two additional tensors $\mathbf{x}_{1,acc}$ and $\mathbf{x}_{2,acc}$ which stores the sum of the increments over previous inference runs. The operation is implemented according to the Eqn~\ref{eq:elementwisemult}. This requires two dense-to-sparse multiplications and two dense-to-sparse tensor additions. The mask of the output can be computed using a bitwise OR operation (Eqn~\ref{eq:elementwisemultmask}). After we perform the multiplication, the quantities $\mathbf{x}_{1,acc}$ and $\mathbf{x}_{2,acc}$ are updated according to Eqn~\ref{eq:elementwisemultx1accum}.
\begin{gather}
\label{eq:elementwisemult}
\mathbf{y}_{\uparrow} = (\mathbf{x}_{1,acc} + \mathbf{x}_{1,\uparrow})\,\mathbf{x}_{2,\uparrow} + \mathbf{x}_{2,acc}\,\mathbf{x}_{1,\uparrow} \\
\label{eq:elementwisemultmask}
\mathbf{y}_{m} = \mathbf{x}_{1,m} \mid \mathbf{x}_{2,m}\\
\label{eq:elementwisemultx1accum}
\mathbf{x}_{1/2,acc} \gets \mathbf{x}_{1/2,acc} + \mathbf{x}_{1/2,\uparrow} 
\end{gather}

\textbf{Convolution on Sparse Inputs}
\label{sec:conv_impl}
\remove{In recent works, there has been a lot of interest in achieving faster execution on sparse operations. Recent works proposed many specialized hardware and hardware-software co-designed techniques~\cite{torchsparse, spconv, submspconv,minkowski} to execute sparse operations efficiently. However, sparse computations on GPUs and CPUs are inefficient due to irregular accesses to memory. For example, performing sparse operations requires the GPU to perform data-dependent uncoalesced memory accesses and writes that lead to high memory access latencies~\cite{minkowski, torchsparse}. Furthermore, operations on sparse tensors may not efficiently leverage vector intrinsics~\cite{spconv}. Thus, sparse convolution operations achieve speedups over dense convolutions only at very high sparsities~\cite{spconv} (above $99\%$).}
\remove{Along with each increment tensor, each operator takes as input a tensor mask. Each element of the mask indicates whether a region of the input tensor can be skipped for computation as it is entirely comprised of zeros. A mask has the same dimensions as the input increment tensor. }
\remove{On receiving as input an increment tensor and a mask tensors with dimension $\mathbb{R}^{H\times W\times C}$ (i.e., height, weight, and channel), each element of the mask summarizes the sparsity of a smaller section of $\mathbb{R}^{h\times w\times1}$ elements in the input tensor, as depicted in Figure~\ref{fig:tensor_fscp}.}
Our convolution layer receives as input an increment tensor and a mask tensor of size $H\times W$ and $C$ channels. The mask of the input contains information on all the locations in the input where $h\times w$ sections of the tensor are $0$.
When computing convolution over the input with a filter of size $k$, we can skip applying the convolution filter overall $k\times k$ regions within each $h\times w$ section of the image of a particular channel. This skipping of sets of filter computations results in faster convolution.

\subsection{Drift Errors}
\begin{figure}[!htb]
\vspace{-0.2cm}
\centering
    \begin{subfigure}{.23\textwidth}
        \centering
        \includegraphics[width=1\linewidth, trim={0cm, 0cm, 0cm, 1.5cm}, clip]{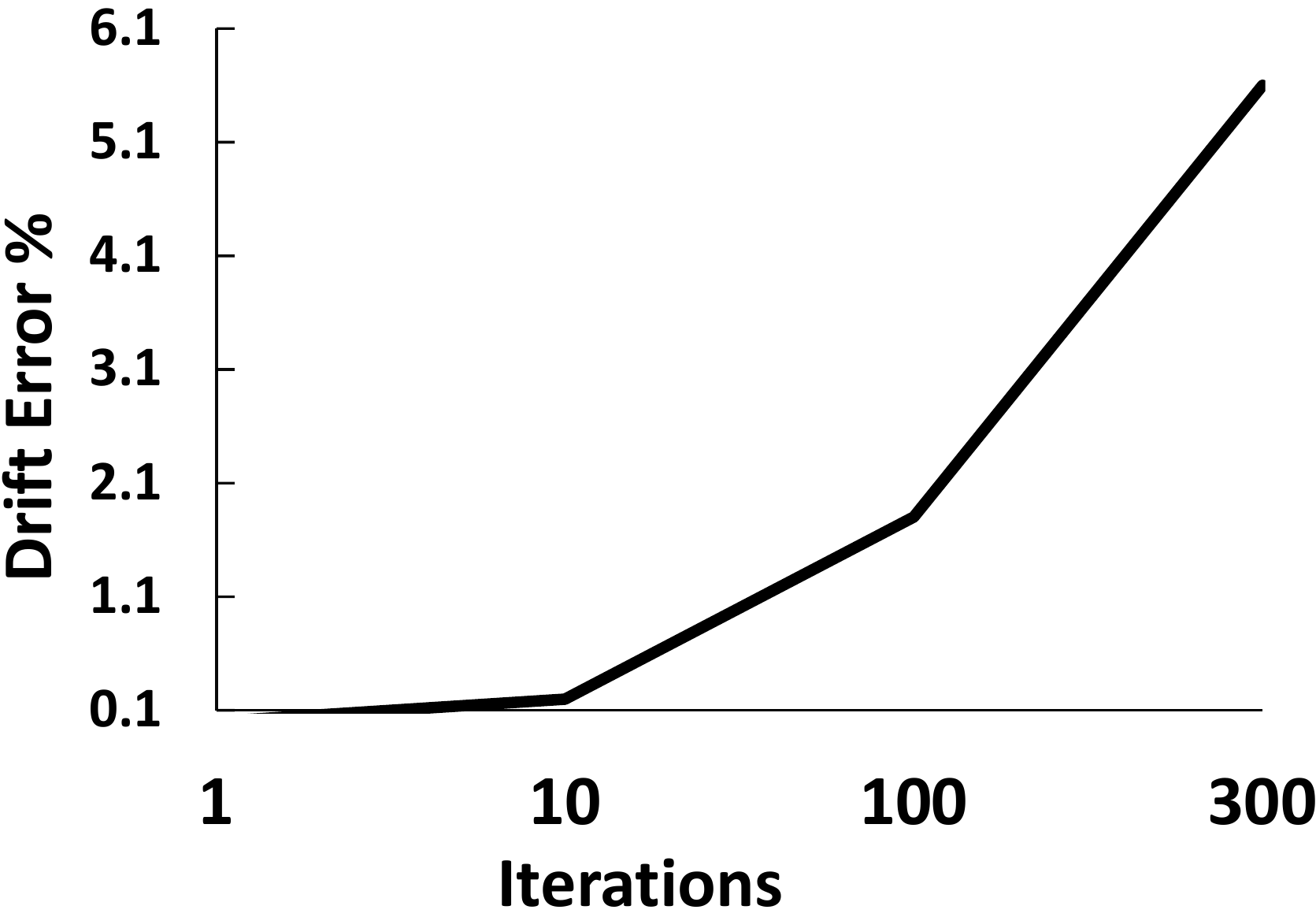}
        \caption{Drift error in max-absolute ouput tensor value of YOLE}
        \label{fig:drift}
    \end{subfigure}
    \begin{subfigure}{.23\textwidth}
        \centering
        \includegraphics[width=1.1\linewidth, trim={0.2cm, .2cm, .1cm, .2cm}, clip]{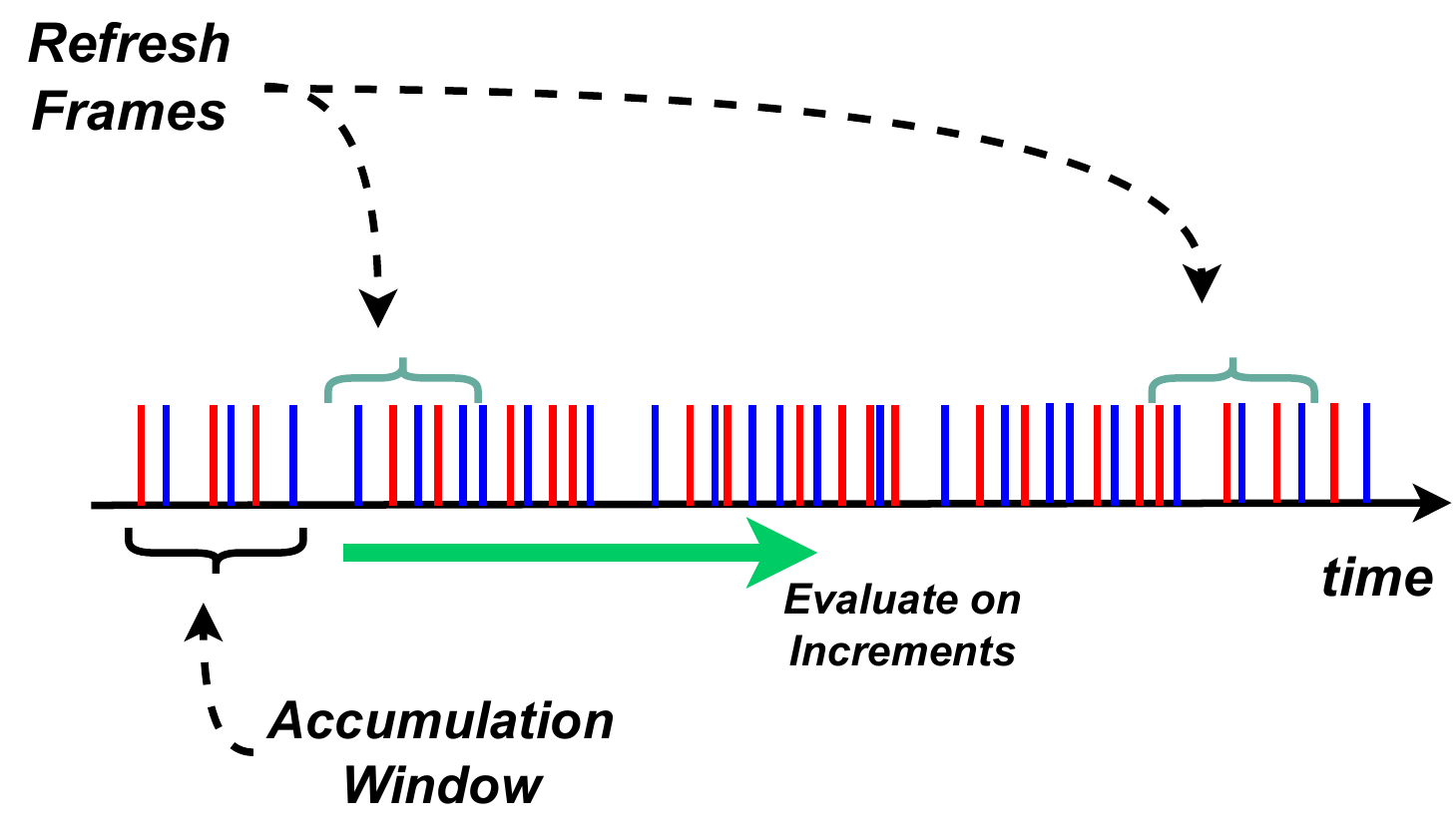}
        \caption{Refresh operation is run at every $N$ inferences}
        \label{fig:refresh}
    \end{subfigure}
    \caption{Drift error in inference output}
    \vspace{-0.6cm}
\end{figure}

Repeated inferences on incremental inputs make \X{} prone to drift errors. We perform an experiment and measure the difference between the output of the incremental and the regular version of a YOLE~\cite{yole} CNN fine-tuned on N-Caltech~\cite{ncaltech} dataset. Figure~\ref{fig:drift} shows the error in the element with the maximum absolute difference between the real output and the output of the network computed with increments. We see that as the number of iterations increases, the percentage of the element with the maximum absolute difference compounds over time. To address this problem, we re-initialize each accumulated tensor ($x_{acc}$ in activations and pointwise multiplications in Section~\ref{sec:incrops}) by performing a regular forward pass. We call this run a refresh step. We run this step every N inferences to reduce the address drift errors, as demonstrated in Figure~\ref{fig:refresh}.

\section{EXPERIMENTS}
We perform all our experiments on a desktop with an Intel 11700K with an RTX 3060 GPU. We evaluate \X{} on three computer vision applications that use different CNN architectures: real-time monocular depth estimation, optical flow estimation and object recognition. We insert our sparsification layer ahead of each convolution operation. We evaluate \X{} for depth estimation tasks on two neural network architectures, a Conv-LSTM UNet architecture based on E2Depth~\cite{rpg_e2depth} and a Bimodal ConvGRU based on RAMNET~\cite{ramnet} on the DENSE~\cite{dense} dataset. These networks use a spatio-temporal voxel grid to encode event inputs into input tensors. \draftmodadd{The DENSE dataset is synthetically produced using an event camera with a resolution of 346x260 and a mean readout rate of 24 kHz.} For object recognition, we evaluate \X{} on N-Caltech101~\cite{ncaltech} and NCars~\cite{hats} datasets with the YOLE CNN~\cite{yole}. \draftmodadd{These datasets are produced with an event camera of resolution of 240x180 and a mean event readout rate of 2.7 kHz.} \X{} for optical flow estimation is evaluated on different neural network architectures: a recurrent convolution-UNet architecture~\cite{event_flow}, UNet architecture based on EVFlowNet~\cite{evflownet}, and a simple sequence of convolutions based on FireFlowNet~\cite{fireflownet} and our delayed UNet architecture on the MVSEC dataset~\cite{mvsec}.\draftmodadd{ The MVSEC dataset consists of many sequences of outdoor and indoor scenes captured using an event camera with a resolution of 346x260 and a mean readout rate of 185-270 kHz.}\draftmodadd{ In all of our experiments we have considered the time window to capture events at 50ms long, and shifting 1ms forward in time.}

\subsection{Inference Latency and Required MFLOPs}
\label{sec:inference_latency}
From Table~\ref{tab:de_results}, we observe that \X{} provides a speedup of up to $1.6\times$ on depth estimation task by leveraging the sparsity in the input increments. Furthermore, we observe a significant reduction in the number of floating-point operations required in the forward pass by as much as $97\%$.  \X{} achieves a similar log-RMSE error compared to the original network. On the object recognition task (Table~\ref{tab:objdet}), we observe a faster inference latency of about  $1.1\times$, resulting from a large reduction in the number of floating-point operations by about $89\%$. We observe no change in accuracy on using \X{} compared to the baseline model. For the optical flow task, we demonstrated a significant reduction in floating-point operations in the forward pass, which reduced over 90\% of the required operations. However, \X{} implementation sees a slight slowdown compared to the PyTorch implementation for RecEVFlowNet.  For optical flow task, we compare the average endpoint error (AEE) (Table~\ref{tab:aee}) of the baseline and the models using \X{}. \draftmod{We observe that our delayed model achieves similar accuracy compared with EVFlowNet.} 
\draftmodadd{Despite the large reduction of floating point operations, the improvement in latency is much smaller in comparison. The reason for this is that we compare our hand-written kernel implementations with highly-optimized cuDNN implementations for dense tensor computations, which invoke specialized optimization techniques (device-specific intrinsics) to achieve better performance. As a number of these techniques are not directly accessible to the user and require expert domain knowledge, we operate with a weaker baseline implementation.}

\begin{table}[htb!]
\centering
\begin{tabular}{|c|c|c|c|c|}

\hline
\textbf{Network} & \textbf{Latency} & \textbf{MFLOPs} & \textbf{RMSE} \\
 &  &  & \textbf{Log error} \\
\hline
         ConvLSTM-UNet~\cite{rpg_e2depth}   &  18ms  & 1023.9 & 0.74 \\
         Bi-ConvGRU-UNet~\cite{ramnet} &    24ms  & 2322.0  & 0.61 \\
        \textbf{\X{}}ConvLSTM-UNet   &      12ms  & 44.1 & 0.77 \\
        \textbf{\X{}}-BiConvGRU       &     15ms  & 54.5 & 0.74 \\
\hline

\end{tabular}

\caption{Evaluation of \X{} for depth estimation on the DENSE town-007~\cite{dense} dataset}
\label{tab:de_results}
\vspace{-0.5cm}

\end{table}

\begin{table}[htb!]
\centering
\begin{tabular}{|c|c|c|c|c|c|}

\hline
\textbf{Network} & \textbf{Latency} & \textbf{MFLOPs} & \textbf{Accuracy-}         & \textbf{Accuracy-} \\
                 &                  &                 & N-Caltech & N-Cars \\

\hline
YOLE~\cite{yole} &   26ms           & 58.9            & 69.5\%                    & 92.4\%  \\
\textbf{\X{}}-YOLE & 24.6ms         & 6.8             & 69.5\%                    & 92\%    \\
\hline

\end{tabular}

\caption{Evaluation of \X{} for object recognition on N-Caltech~\cite{ncaltech} and N-Cars~\cite{hats} datasets}
\label{tab:objdet}
\vspace{-0.7cm}
\end{table}

\begin{table*}[htb!]
\vspace{.1cm}
\centering
\begin{tabular}{|c|c|c|c|c|c|c|}

\hline
\textbf{Network Architecture} &  \textbf{Latency} & \textbf{MFLOPs} & \textbf{outdoor day 1} & \textbf{indoor flying 1} & \textbf{indoor flying 2} & \textbf{indoor flying 3} \\
\hline
           RecEVFlowNet~\cite{evflownet}    &    16.6ms  &  818.7 &   0.47        & 0.59              &   1.17        & 0.93 \\
           EVFlowNet~\cite{evflownet} &      9.5ms   & 364 &  0.85             &    1.16               &    2.00           &  1.12 \\
           FireFlowNet~\cite{fireflownet}    &      4.4ms  &  122.4 &  1.02         &   1.37            &   2.24        & 2.00 \\
           \textbf{Delayed Multires-UNet} &   8.1ms    & 416.7  &   0.81            &  1.19                 &   2.18        & 1.95  \\

                       \textbf{\X{}}RecEVFlowNet &   17.6ms  & 72.0 &  0.53         & 0.66              & 1.23          & 0.98  \\
                       \textbf{\X{}}EVFlowNet     & 8.91ms     & 62.5 & 0.89              &  1.37                 & 2.24          & 2.00 \\
                       \textbf{\X{}}FireFlowNet   &  3.8ms    & 15.9  & 1.02       & 1.3             &   2.1        &  2.00 \\
                       \textbf{\X{}}Delayed-Multires-UNet  &   6.6ms   & 51.9 &   0.75            &  1.2                 &   2.11        & 1.95 \\ 
\hline
\end{tabular}

\caption{Evaluation of \X{} for the optical flow task on MVSEC~\cite{mvsec} sequences}
\label{table:inference_latency}
\label{tab:aee}
\vspace{-0.5cm}
\end{table*}

\subsection{Ablation Study}
\begin{figure}[!htb]
    \centering
    \includegraphics[width=0.8\linewidth]{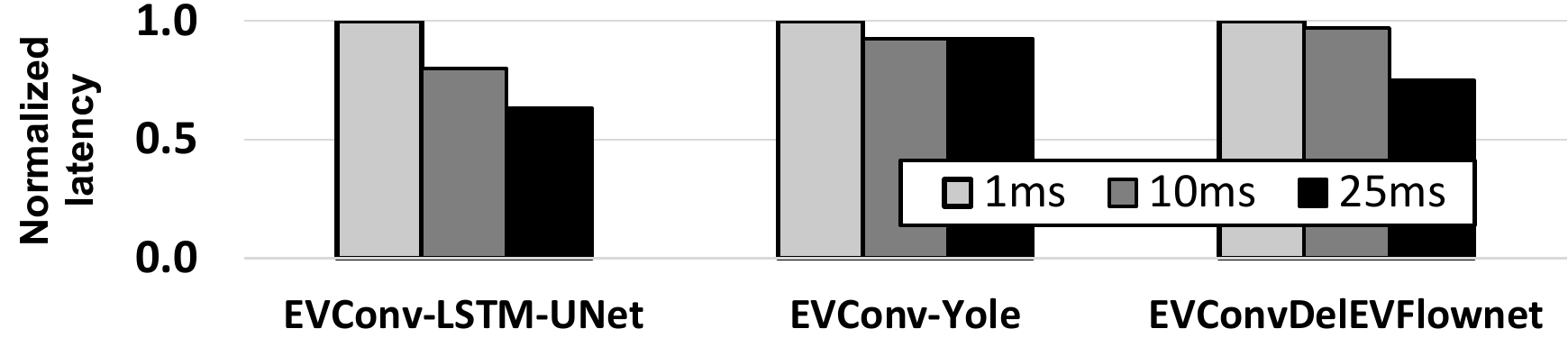}
    \caption{\draftmodadd{Normalized latency on varying the inference timestep}}
    \label{fig:results_timestep}
    \vspace{-0.3cm}
\end{figure}

\vspace{-0.3cm}
\draftmodadd{\subsubsection{Changing the inference timestep}
Our framework performs inference over a time window of events. This window is shifted forwards in time each time when performing inference. The inference speed depends on the sparsity of the increment tensor observed, which depends on the inference rate. We present here the effect of varying the length of the inference timestep on inference speed in Figure~\ref{fig:results_timestep}. }

\subsubsection{Changing the threshold parameter}
\begin{figure}[!htb]
\vspace{-0.2cm}
    \centering
    \includegraphics[width=0.8\linewidth]{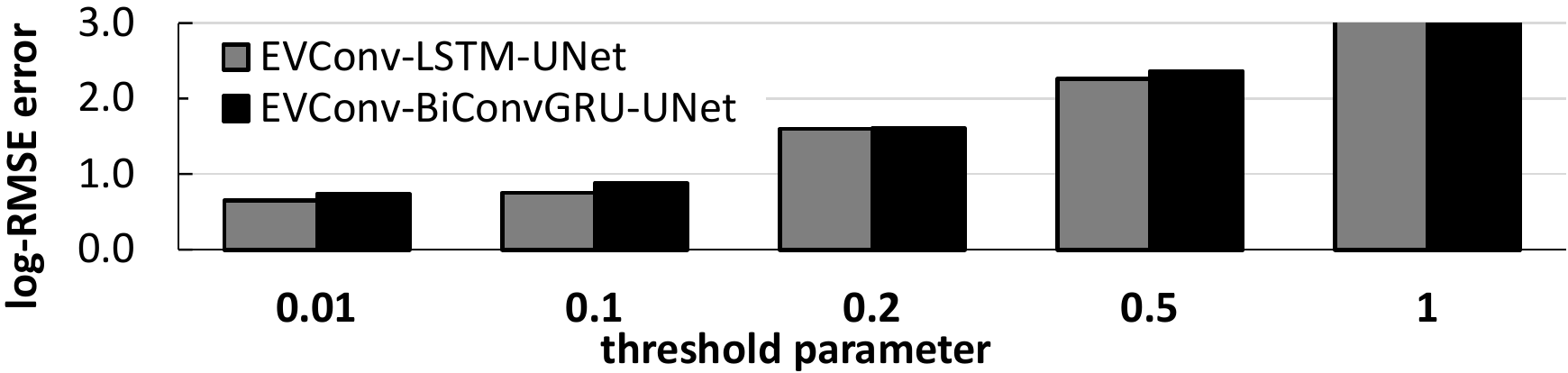}
    \caption{log-RMSE error on changing the threshold parameter}
    \label{fig:rmselog}
\end{figure}

\begin{figure}[!htb]
    \vspace{-0.2cm}
    \centering
    \begin{subfigure}{.23\textwidth}
        \centering
        \includegraphics[width=0.9\linewidth, trim={0 1.1cm 0 1.1cm},clip]{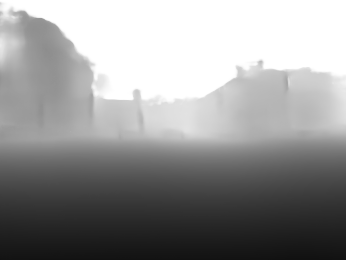}
        \caption{$t_p=0.01$}
        \label{fig:k01}
    \end{subfigure}
    ~
    \begin{subfigure}{.23\textwidth}
        \centering
        \includegraphics[width=0.9\linewidth, trim={0 1.1cm 0 1.1cm},clip]{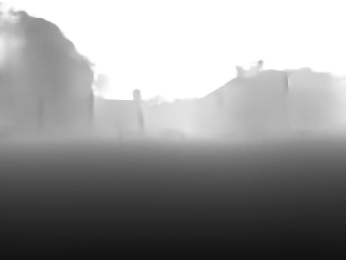}
        \caption{$t_p=0.05$}
        \label{fig:k05}
    \end{subfigure}
    ~
    \begin{subfigure}{.23\textwidth}
        \centering
        \includegraphics[width=0.9\linewidth, trim={0 1.1cm 0 1.1cm},clip]{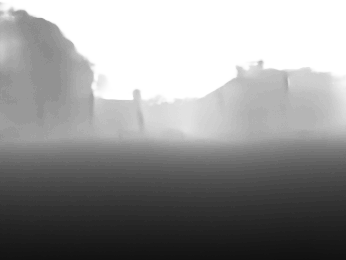}
        \caption{$t_p=0.1$}
        \label{fig:k1}
    \end{subfigure}
    ~
    \begin{subfigure}{.23\textwidth}
        \centering
        \includegraphics[width=0.9\linewidth, trim={0 1.1cm 0 1.1cm},clip]{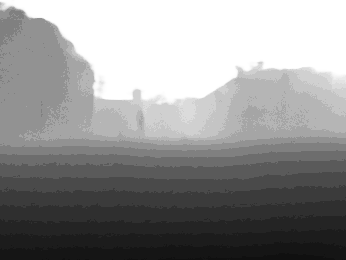}
        \caption{$t_p=0.3$}
        \label{fig:k3}
    \end{subfigure}

    \caption{Depth maps computed by \X{}-ConvLSTM-UNet on a single frame of DENSE town-007 dataset~\cite{dense}.}
    \label{fig:depth_visual}
    \vspace{-0.5cm}
\end{figure}

On depth estimation task, Figure~\ref{fig:rmselog} show how the log-RMSE error changes as the threshold parameter increases. \draftmodadd{Figure~\ref{fig:depth_visual} shows the log-depth estimates of the network at different sparsification threshold $t_p$. We observe that the depthmap is visually similar to baseline.}

\begin{figure}[!htb]
\centering
    \begin{subfigure}{.425\textwidth}
        \centering
        \includegraphics[width=0.85\linewidth]{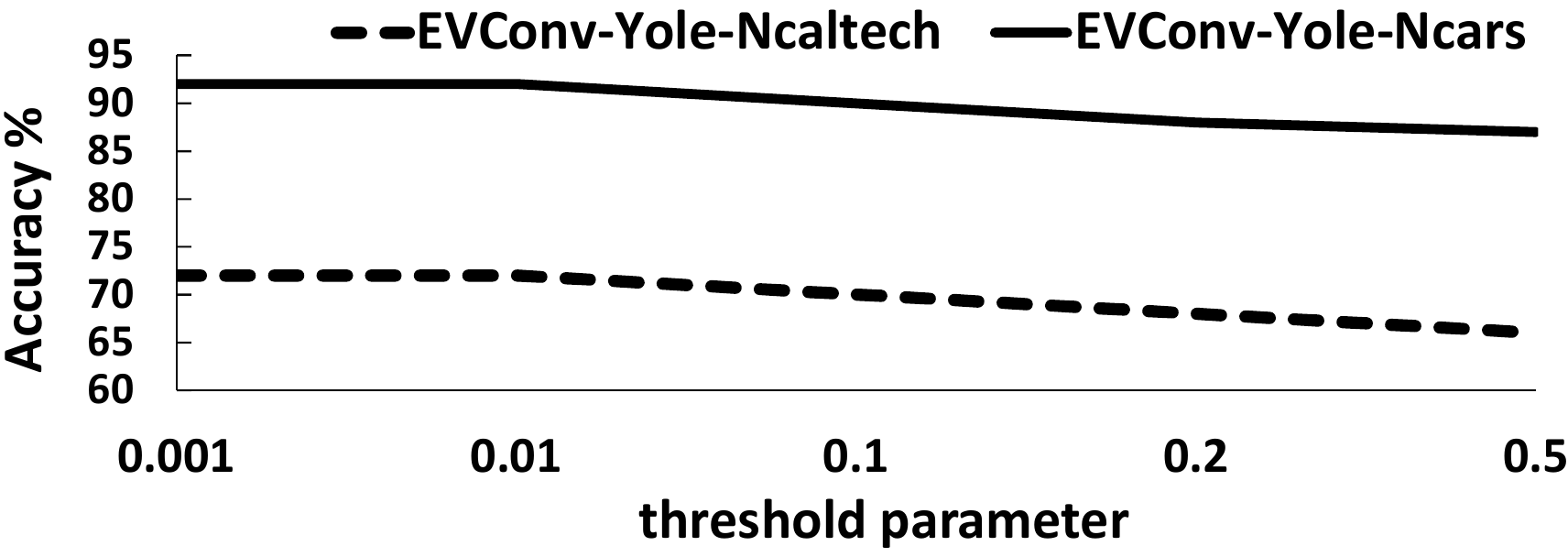}
        \caption{Accuracy: Yole}
        \label{fig:accuracy}
    \end{subfigure}
    \begin{subfigure}{.425\textwidth}
        \centering
        \includegraphics[width=0.9\linewidth]{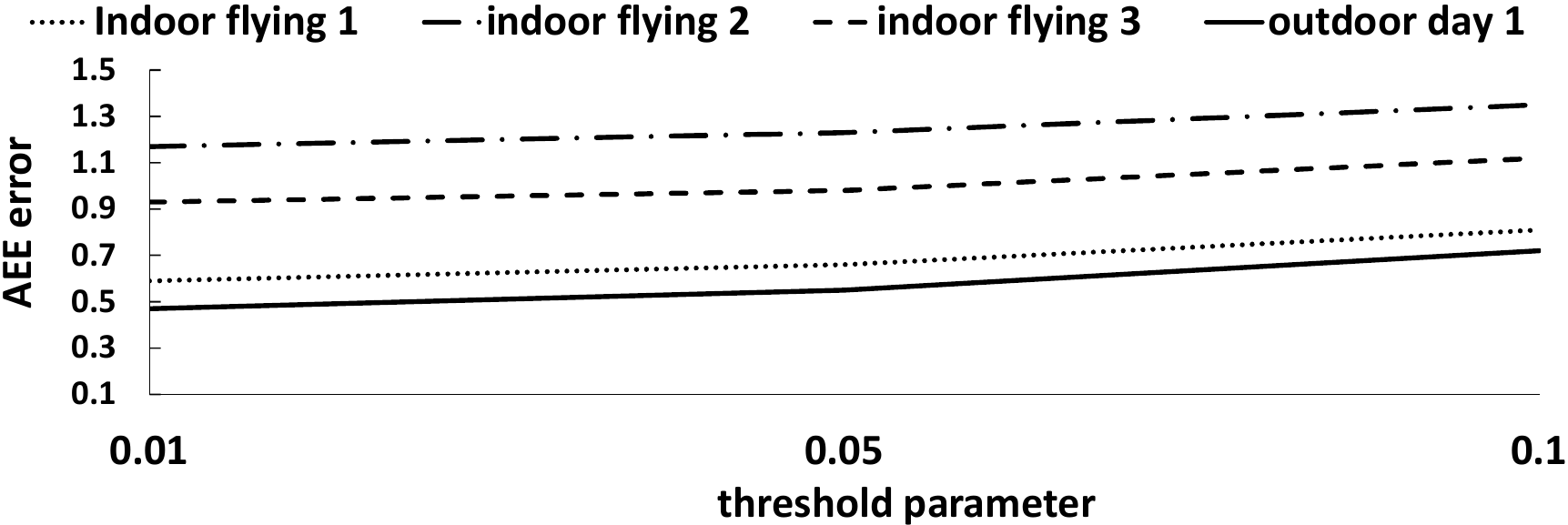}
        \caption{AEE error: RecEVFlowNet}
        \label{fig:aee}
    \end{subfigure}    
    \caption{\X{} on changing the threshold parameter}
    \label{fig:de}
    \vspace{-0.5cm}
\end{figure}

 Figure~\ref{fig:accuracy} shows how the accuracy of the predictions varies with increasing threshold parameter. We see that the accuracy drops marginally (less than $1\%$) on both the N-Caltech and NCars datasets when the threshold parameter is set to $0.1$. Figure~\ref{fig:aee} shows how the Average Endpoint Error (AEE) varies with increasing the threshold parameter on different MVSEC dataset sequences. We see only minor degradation of the AEE error compared to the original network when the threshold parameter is set to $0.1$.

\begin{figure}[!htb]
    \centering

    \begin{subfigure}{0.45\textwidth}
        \centering
        \includegraphics[width=.85\linewidth]{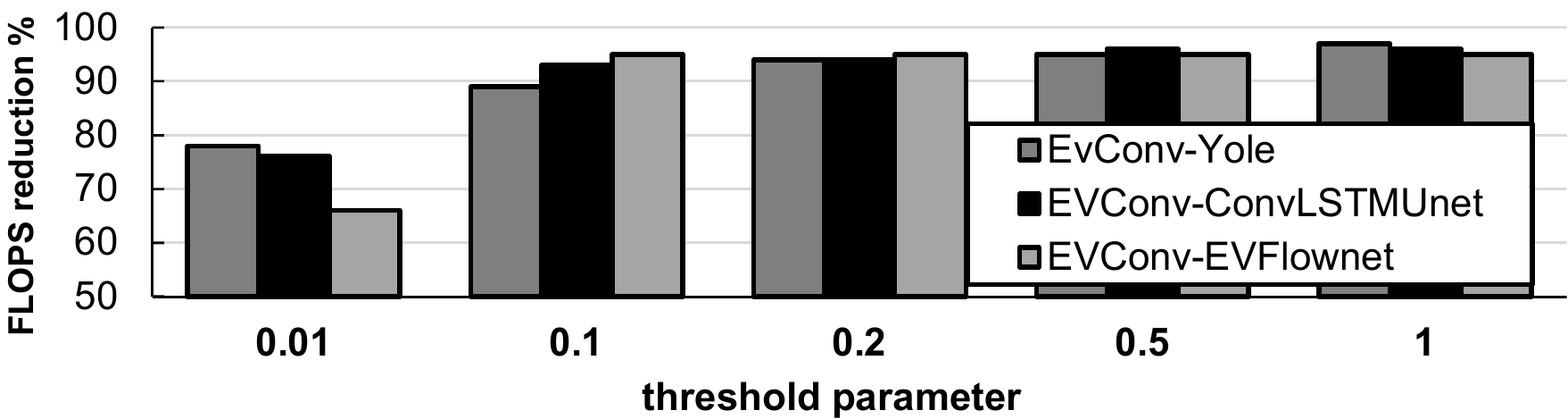}

        \caption{\draftmodadd{\% of FLOPs reduction}}
        \label{fig:flopsred}
    \end{subfigure}
    \begin{subfigure}{0.45\textwidth}
        \centering
        \includegraphics[width=.85\linewidth]{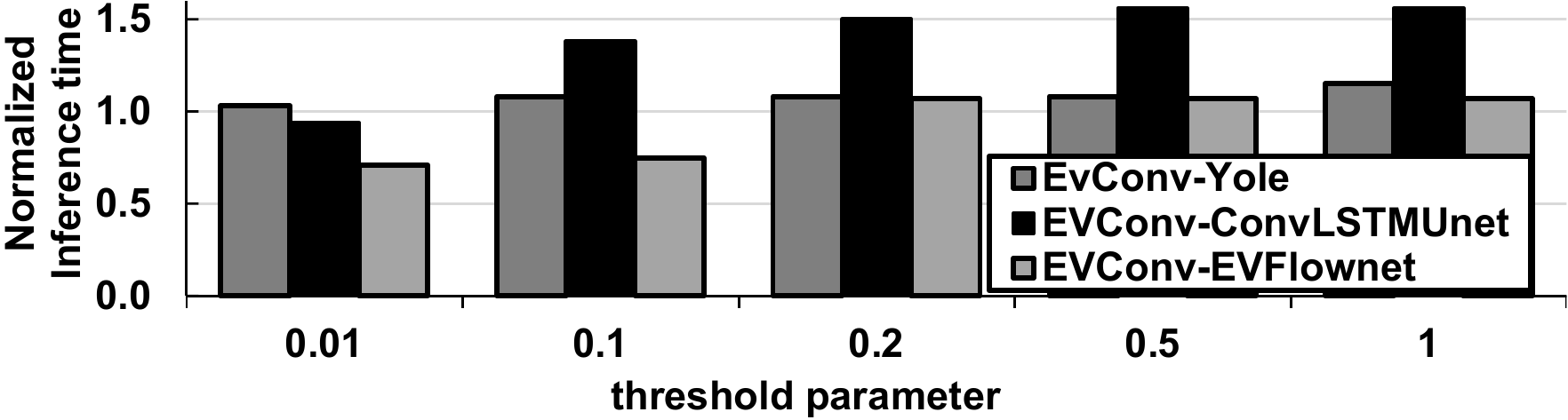}        \caption{\draftmodadd{Normalized inference latency w.r.t cudnn baseline}}
        \label{fig:inflatenct}
    \end{subfigure}

    \caption{\draftmodadd{Inference speed and \% reduction of FLOPs on changing the threshold parameter}}
    \label{fig:visual}
    \vspace{-0.5cm}

\end{figure}

Figure~\ref{fig:flopsred} shows a percentage of the number of FLOPs required for the forward pass compared to baseline on changing the threshold parameter.  
On the depth estimation task, we find that on setting the threshold parameter at $0.1$, we observe a nearly over $97\%$ reduction in the raw number of floating-point operations required. and a $90\%$ reduction for the object detection task.
Figure~\ref{fig:inflatenct} depicts the normalized change in the inference latency on increasing the threshold parameters compared to the baseline. For the depth estimation task, we see a $1.6\times$ and $1.5\times$ speedup in inference latency over our implementation with virtually no change in the RMSE-log error in depth when the threshold parameter is set to $0.1$.
For the optical flow task, despite a significant reduction in floating-point operations, we observe that our best inference latency is slower by about $1.05\times$.
For the object detection task, we see an $8\%$ reduction in the inference latency at the threshold parameter set to 0.1.

\subsection{Comparison to prior works}
\draftmodadd{DeltaCNN~\cite{deltacnn} leverages the sparsity of the difference in consecutive input tensors for faster inference in videos. We adapted DeltaCNN for the event-based tasks for comparison. The convolution operation in DeltaCNN can skip computations if all elements of the receptive field of a convolution filter are 0, throughout the channel dimensions. For example, if the receptive field of the convolution is 3x3, 3x3xC channel dimensions are required to be 0. Although high, the sparsity of increments of event camera inputs is very irregular. Our implementation is able to handle irregular sparsity by requiring a tile of dimension 6x6 to be 0 to be able to skip computations (as indicated in Section VI). Additionally, DeltaCNN skips the cooperative memory fetch of the input tensor when all elements to be loaded are 0. We perform an experiment where we use DeltaCNN for inference on event camera input encodings, shown in Figure 10. On incorporating this memory load optimization, our inference speeds improve further, as seen under the ‘ours + conditioned fetch’ label.
\begin{figure}[!htb]
    \centering
    \includegraphics[width=.88\linewidth]{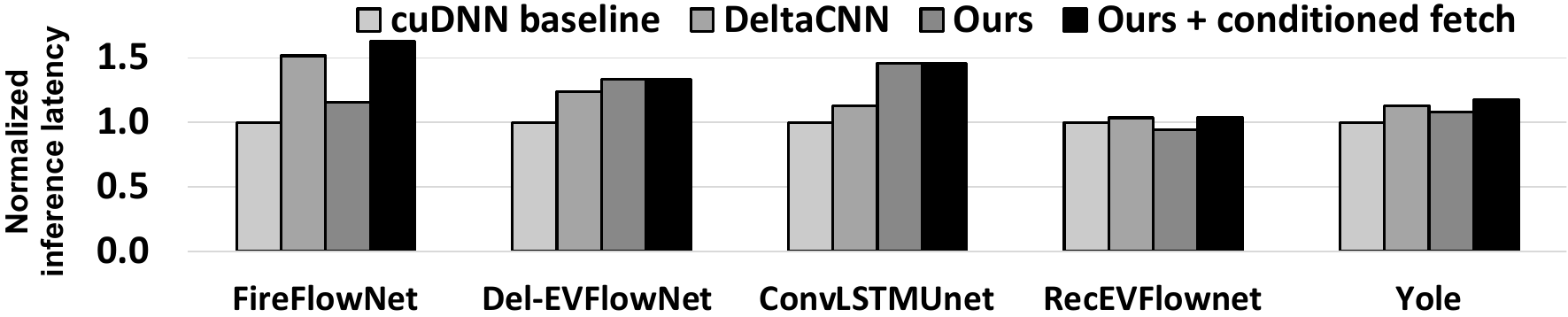}
    \caption{Comparison with DeltaCNN~\cite{deltacnn}}
    \label{fig:deltacnn}
    \vspace{-0.1cm}
\end{figure}

AsyNet~\cite{asynchronous_sparse_cnn} uses submanifold sparse convolution to leverage sparsity in event encodings while AEGNN~\cite{aegnn} uses a graph neural network over event encodings. We compare the inference speed of these works with our work\remove{ in Table~\ref{tab:comparison_other_works}}. We find that AsyNet performs $2.31\times$, and AEGNN performs $2.96\times$ slower inference compared to our CNN implementation due to irregular memory accesses in doing inference with submanifold sparse convolutions and graph neural networks on the GPU.}
\remove{
\begin{table}[htb!]
    \centering
    \begin{tabular}{|c|c|c|c|}
        \hline
\textbf{} &
\textbf{Ours} &
\textbf{AsyNet~\cite{asynchronous_sparse_cnn} } &
\textbf{AEGNN~\cite{aegnn}} \\

        \hline
\textbf{Inference latency} & $1\times$ & $2.31\times$ & $2.96\times$ \\
    \hline
    \end{tabular}
    \caption{Comparison of techniques used in prior works}
    \label{tab:comparison_other_works}
\end{table}
}

\vspace{-0.2cm}

\section{CONCLUSION}

We introduced \X{}, a new approach to achieving fast CNN inference on event camera inputs \draftmod{to enable high-speed robotics perception such as high-speed motion estimation, event-based object detection, object tracking, etc.} We observe that at high inference rates, the difference between consecutive inputs to the CNN is small. \X{} thus performs inference on the difference between consecutive inputs, or \emph{increments}, rather than the event camera stream itself. \X{} leverages the sparsity of the increments to effectively accelerate inference by reducing the number of floating points required. \X{} is designed to retain the sparsity of the inputs to all layers of the CNN without sacrificing the accuracy of the network. \X{} is able to reduce the floating operations by up to $98\%$ and provide up to $1.6\times$ faster inference speed on depth estimation, optical flow, and object recognition.





{
\small

\bibliographystyle{plain}
\bibliography{refs_compressed}

}

\end{document}